\pdfoutput=1

\documentclass[11pt]{article}

\usepackage[]{EACL2023}

\usepackage{times}
\usepackage{latexsym}

\usepackage[T1]{fontenc}

\usepackage[utf8]{inputenc}

\usepackage{microtype}

\usepackage{inconsolata}

\usepackage{graphicx}
\usepackage{booktabs}
\usepackage{multirow}
\usepackage[disable]{todonotes}
\usepackage{amsmath}
\usepackage[normalem]{ulem}
\usepackage{caption}
\usepackage{subcaption}
\usepackage{mathtools}

\usepackage{amsmath}
\usepackage[normalem]{ulem}
\usepackage{booktabs}
\usepackage{multirow}
\usepackage{cleveref}

\newcommand{\mbert}[0]{\textsc{mBERT}}
\newcommand{\xtreme}[0]{\textsc{XTREME}}

\newcommand{\ceunbalanced}[0]{\textit{HL}}
\newcommand{\cebalanced}[0]{\textit{HL balanced}}

\newcommand{\highres}[0]{high-resource}
\newcommand{\lowres}[0]{low-resource}

\newcommand{\inlang}[0]{\textsc{In-Lang}}
\newcommand{\zeroshot}[0]{\textsc{Zero-Shot}}

\newcommand{\draftcomment}[3]{{\textcolor{#3}{[#1]#2}}}
\renewcommand{\draftcomment}[3]{}  

\newcommand{\gabi}[1]{\draftcomment{#1}{\textsc{gabi}}{red}}
\newcommand{\gabis}[1]{\gabi{#1}}

\newcommand{\dan}[1]{\draftcomment{#1}{\textsc{dan}}{blue}}

\definecolor{bottlegreen}{rgb}{0.0,0.42,0.31}
\definecolor{donorred}{RGB}{228.,116.,95.}
\definecolor{reciepientblue}{RGB}{0,152,251}
\newcommand{\tomasz}[1]{\draftcomment{#1}{\textsc{tomasz}}{bottlegreen}}

\title{You Can Have Your Data and Balance It Too:\\
Towards Balanced and Efficient Multilingual Models}

\author{Tomasz Limisiewicz$^{\spadesuit}$\thanks{$\;\;$Equal contribution. The order was decided by a coin toss.}   \thanks{$\;\;$Work done while visiting the Hebrew University.} \qquad
  Dan Malkin$^{\diamondsuit}$\footnotemark[1] \qquad
  Gabriel Stanovsky$^{\diamondsuit}$ \\
  $^{\diamondsuit}\;$School of Computer Science, The Hebrew University of Jerusalem \\
  $^{\spadesuit}\;$Faculty of Mathematics and Physics, Charles University in Prague
 \\
  \texttt{\{dan.malkinhueb,gabriel.stanovsky\}@mail.huji.ac.il} \\
  \texttt{limisiewicz@ufal.mff.cuni.cz}}

\begin{document}
\maketitle
\begin{abstract}
Multilingual models have been widely used for cross-lingual transfer to low-resource languages. However, the performance on these languages is hindered by their under-representation in the pretraining data. To alleviate this problem, we propose a novel multilingual training technique based on teacher-student knowledge distillation. 
In this setting, we utilize monolingual teacher models optimized for their language. We use those teachers along with balanced (sub-sampled) data to distill the teachers' knowledge into a single multilingual student.
Our method outperforms standard training methods in low-resource languages and retains performance on high-resource languages.
\footnote{We will make all of our code and resources publicly available.}

\end{abstract}

\section{Introduction}
\label{sec:intro}

\begin{figure}[!t]
    \centering
    \includegraphics[width=0.87\linewidth]{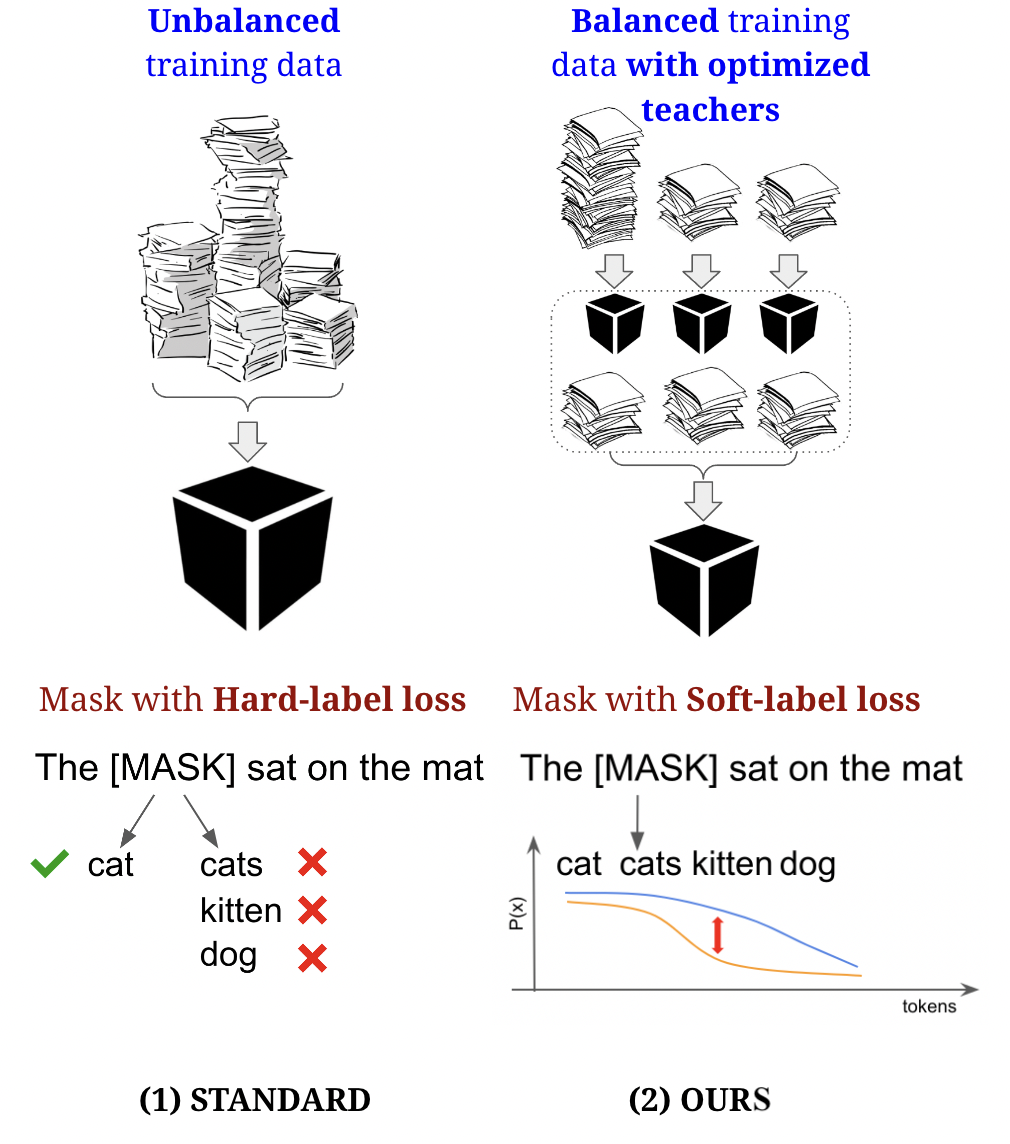}
    \caption{
    We train a student for multilingual language modeling using a collection of teachers optimized for each of the target languages and multilingual data sub-sampled to the data size of the lowest resource language. 
    Our approach achieves a better trade-off in performance between high- and low-resource languages. 
     }
    \label{fig:method}
\end{figure}

While multilingual language models have been gaining popularity, largely thanks to their cross-lingual transfer ability, their performance has been shown to be skewed toward languages with abundant data~\citep{joshi-etal-2020-state,wu2020all}.
Introducing language models that better incorporate diverse and low-resource languages can increase accessibility to NLP technologies in these languages and help improve cross-lingual transfer~\citep{malkin2022balanced}. 

In this work, we address two research questions. First, we ask if we can we improve performance on low-resource languages without hurting it on high-resource ones? Second, does a better trade-off between high- and low-resource languages improve cross-lingual transfer?

To answer these two questions, we distill multiple monolingual teacher models optimized for various languages into a single multilingual student model, using a small balanced multilingual dataset (Figure~\ref{fig:method}).
Our experiments show that this allows taking advantage of data in high-resource languages while avoiding under-fitting \lowres{} languages.

\section{Background: Soft Vs. Hard Labels}
\label{sec:bg}
We compare two alternatives for the masked LM loss functions: the original loss used for masked language modeling, i.e., \textit{hard labeling} and \textit{soft labeling} as defined in  \citet{sanh2019distilbert}: 

\noindent (1) \textit{hard labeling}, which takes into account a single gold masked token in a sentence, $y_{gold}$, and evaluates the model's prediction for this word, i.e., standard cross-entropy loss: 
\begin{equation}
    \mathcal{L}_{HARD} 
    = - \log (P(y_{gold}))
    \label{eqn:hl-loss}
\end{equation}

\noindent (2) \textit{soft labeling}, which allows for multiple valid candidates using the output distribution of an oracle (or a strong LM) $\hat{M_l}$ as a soft label:
\begin{equation}
    \mathcal{L}_{SOFT} = - \sum_{y \in V} P_{\hat{M_l}}(y) \log \frac{P(y)}{P_{\hat{M_l}}(y)}
    \label{eqn:sl-loss}
\end{equation}

Where $y$ denotes tokens in the model's vocabulary $V$. Please note that $\mathcal{L}_{SOFT}$ is also equivalent to a KL-divergence between oracle and predicted distributions.

In the following sections, we will explain how soft labeling allows us to distill multiple teachers into a single multilingual student while accounting for balanced performance in high- and low-resource languages.

\section{Teacher-Student Distillation for Multilingual Language Models}
\label{sec:method}

We train a multilingual student using the masked-language modeling objective and a collection of monolingual teachers optimized for each student's language. All models share one multilingual vocabulary. Sharing vocabulary was necessary to apply our \textit{soft labeling} loss, which requires that the student's and teacher's probability space (in the case of language models: vocabularies) are the same.\footnote{Please refer to Section~\ref{sec:limitations}, ``Teacher model availability'' for discussion about vocabulary sharing across monolingual models.}

To avoid under-fitting low-resource languages, we naively balance the students' training data by truncating data in all target languages to the data size of the lowest resource language. 
To make the most out of high-resource languages, we rely on 
soft labeling.
For a mask in a given language, we use the \emph{high-resource} language-specific teacher's distribution over the mask and use it as the oracle $\hat{M}_l$ in Equation~\ref{eqn:sl-loss} as a soft label. Our intuition is that this allows the student to gain the broader teachers' knowledge in its language and thus compensate for the sub-sampled data size. Figure \ref{fig:method} provides a visual scheme for this approach.

Formally, given a set of languages $L=\{l_1, l_2, ..., l_K\}$, their corresponding teachers $T_{l_1}, T_{l_2}, ..., T_{l_K}$, and their data $D=\{D_1, D_2, ..., D_K\}$ we teach the student model using the $K$ teachers (which are trained for each of the languages). For student training, we truncate the data size of all languages in $D$ to the smallest dataset size ($min(|D_1|, |D_2|, ..., |D_K|)$).


\label{sec:experimental_setting}

\begin{table}[t!]
\centering
\small
\begin{tabular}{@{}ccc@{}}
\toprule
 \textbf{Size [characters]} &
 \textbf{Shared Script} &
 \textbf{Diverse Script}
   \\ \midrule
100M & English    &  Russian  \\
100M & German     &  German   \\
50M  & Spanish    &  Korean   \\ \midrule
30M  & Hungarian  &  Greek    \\ 
20M  & Vietnamese &  Hindi    \\
10M  & Turkish    &  Telugu   \\
10M  & Basque     &  Urdu     \\ \bottomrule
\end{tabular}%
\caption{
Pre-training datasets for each language (in millions of characters) sampled from Wikipedia for 
\highres{} (top) and \lowres{} (bottom) languages.
Some of the selected \lowres{} languages are actually widely spoken. They were chosen because of relatively smaller Wikipedia sizes (as shown in Appendix~\ref{sec:data-details}).
}
\label{tab:lang}
\end{table}

\paragraph{Data selection and processing.}
\label{sec:data_selection}
We collect pretraining data from Wikipedia,\footnote{Obtained and cleaned using wikiextractor~\citep{Wikiextractor2015}. We chose Wikipedia as it consists of roughly similar encyclopedic domains across languages and is widely used for training PLMs~\citep{devlin-etal-2019-bert}.} aiming to capture a diverse set of high and low-resource languages, as summarized in Table~\ref{tab:lang}.
We subsample the corpora by randomly choosing sentences from each language's full corpus.
We designate \highres{} languages as ones with 50 or 100 million characters in their corpus after sampling, while \lowres{} languages' corpora consist of 10, 20, and 30 million characters.

Throughout our experiments, we compare 7 languages that share the Latin script versus 7 languages with varying scripts, as the script was found to be an essential factor for multilingual performance~\citep{wang2019cross,muller2020being,malkin2022balanced}. We include German in both sets (as one of 7 languages), to compare its performance in both settings.

\paragraph{Models' Architecture and Hyper-parameters.}
\label{sec:hyperparameters}
Each of our models comprises of 6 hidden layers and 4 attention heads, an MLM task head. The embedding dimension is 512 and sentences were truncated to 128 tokens. In total, our models consist of 51193168 parameters. We train a single uncased wordpiece tokenizer ~\citep{wu2016google} on the 100mb splits of 15 languages.\footnote{13 languages presented in Table~\ref{tab:lang} with Hebrew and Lithuanian that were added for future experiments.} 
\dan{The shared vocabulary tokenizer allowed us to distill the teachers' knowledge easily in our controlled experiments}\tomasz{I already addressed reason for sharing vocab in the first paragraph of section 3 above.}
Before tokenization, we strip accents for all languages except Korean.

We train all models for 10 epochs, with a batch size of 8. We used linear decay of the learning rate with the initial value of 5e-5.
Exact configurations and parameters are available in our code. 

\section{Experiments}
\label{sec:experiments}
We validate our method using two experiments. First, we ascertain that our method indeed improves performance for low-resource languages while maintaining performance for high-resource languages. This is done by comparing the performance of our approach in masked language modeling with two multilingual baselines. Second, we show that our method is competitive for downstream tasks and cross-lingual transfer by probing the pre-trained models for POS and NER tagging.

\paragraph{Multilingual modeling.} We evaluate masked language modeling performance on monolingual test sets by measuring \emph{mean reciprocal rank} (MRR).
Since the performance of multilingual models is often compared to the performances of monolingual baselines, we report the \emph{average performance difference} between a multilingual model and the monolingual models trained on the same set of respective languages. 

\paragraph{Downstream probing.} 
We use the models trained in the previous experiment and train a probe,\footnote{} keeping the base model parameters frozen, to predict part-of-speech tagging (POS) and name entity recognition (NER), as provided respectively by universal dependencies~\cite{nivre-etal-2020-ud} and the \xtreme{} benchmark~\cite{hu2020xtreme}.\footnote{See Section~\ref{sec:app-ds-data} in the Appendix for more information.}
We chose those two tasks because they commonly appear in NLP pipelines ~\citep{manning-etal-2014-stanford, spacy2}.
We measure the models' performance in two cases: when the training and test datasets are in the same language (denoted \inlang{}) and when a probe trained for a language $l_1$ is tested on another one $l_2$ (denoted \zeroshot{}). 
As noted by \citet{hu2020xtreme}, zero-shot evaluation is a good measure of a model's cross-lingual transfer.
We use probing because it offers a good insight into the representation learned by the model \cite{belinkov-2022-probing}.



\paragraph{Baselines.}
We compare the students' performance to multilingual models trained with \emph{hard labels}, on the same data and languages as the student and its teachers. One such model was trained on all the available data in each language to examine the extent of under-fitting low-resource languages, denoted \textbf{\ceunbalanced{}}. Additionally, to measure how much our student gains from its teacher's knowledge, we train another model on the corpora constrained to the size of the least resourceful language using the standard \emph{hard labels}, denoted \textbf{\cebalanced{}}.

\paragraph{Experimental Setup}
\label{sec:model-architecture}
Each teacher is a monolingual model trained with \emph{hard labels}. The teachers are trained on the entire training corpus available in their language.
In a student model, we distill the knowledge of multiple monolingual teachers into a multilingual student using \emph{soft labels}, as described above. The distillation into the student is performed on groups of shared and diverse script languages. The data is constrained to 10 million characters for each language.
All our models are trained using default BERT hyper-parameters detailed in Section~\ref{sec:hyperparameters}.

\section{Results}
\label{sec:results}
\begin{figure}[!t]
    \centering
    \includegraphics[width=\linewidth]{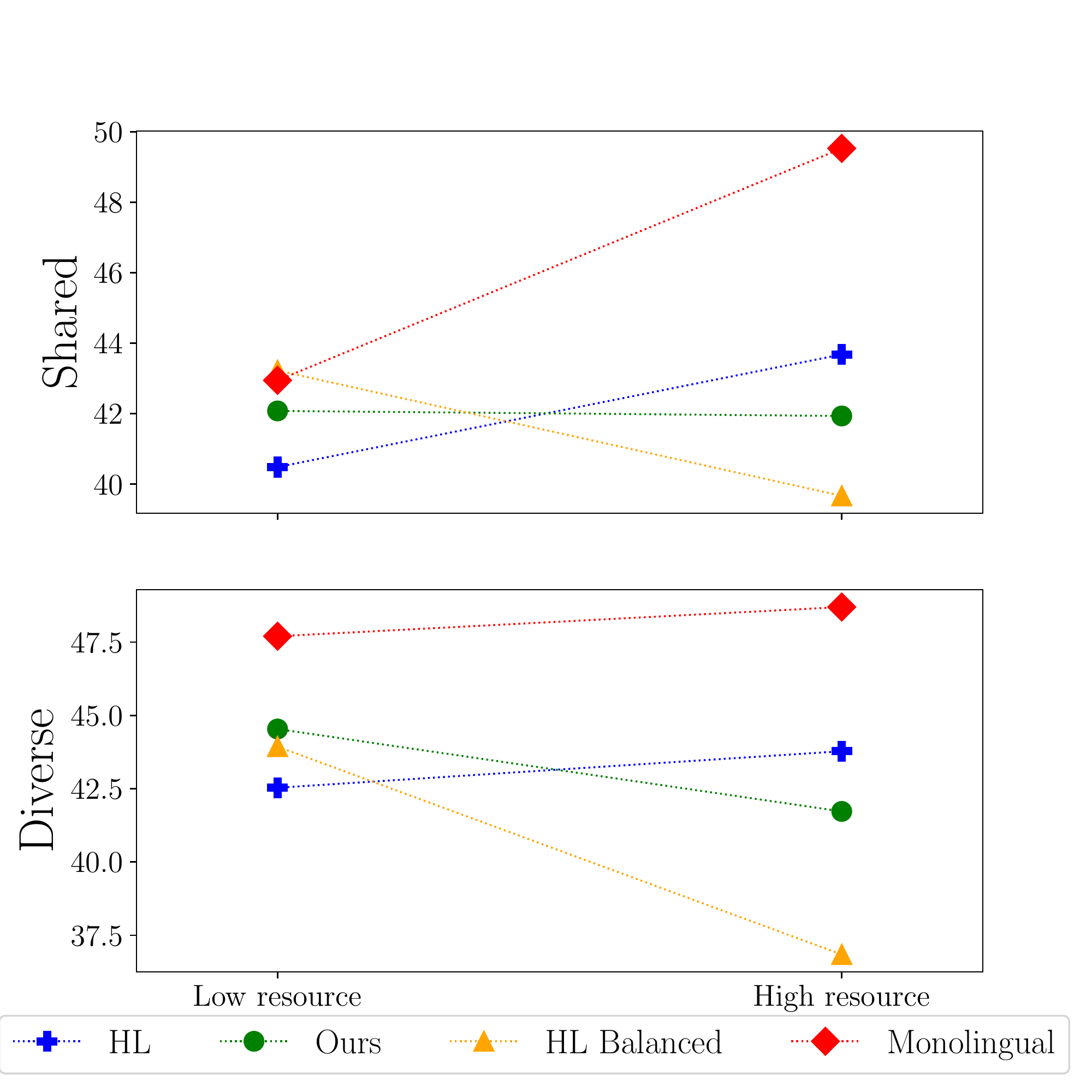}
    \caption{Our balanced teacher-student approach using soft labels presents the overall best combination for low and high-resource languages among multilingual models.
    This figure presents average MRR results in masked language modeling for both low- and high-resource languages. Results are reported for a Latin-script language set (Shared) and a set with diverse scripts (Diverse).
    \gabis{Consider changing ``Shared'' to ``Latin Script'', ``Diverse'' to ``Multiple Scripts''}
    }
    \label{fig:lm-results}
\end{figure}

\begin{table}[t!]
\centering
\small
\resizebox{\columnwidth}{!}{
\begin{tabular}{@{}ccccc@{}}
\toprule
\textbf{Script}                   & \textbf{Lang. Set} & \textbf{HL}   & \textbf{HL Balanced} & \textbf{Ours} \\ \midrule
\multirow{3}{*}{\textbf{Shared}}  & Low-Res.  & -2.5          & \textbf{0.3}         & -0.1          \\
                                  & High-Res. & 
                                  \textbf{-5.8} & -10                  & -7.6          \\
                                  & All       & -3.9          & -4.0                 & \textbf{-3.7} \\ \midrule
\multirow{3}{*}{\textbf{Diverse}} & Low-Res.  & -5.1          & -3.8                 & \textbf{-3.1} \\
                                  & High-Res. & \textbf{-5.0} & -12                  & -7.0          \\
                                  & All       & -5.0          & -7.2                 & \textbf{-4.7} \\ \bottomrule
\end{tabular}
}
\caption{Average difference from monolingual baselines (higher is better) calculated on MRR scores. Our teacher-student model achieves better results overall in both shared and diverse scripts. It is otherwise between the baselines, except for shared script, where it is better for \lowres{}.}
\label{tab:dist_from_mono}
\end{table}
We report the experimental results on our test sets, in three language sets grouped by the amount of data available in pre-training, i.e., \lowres{}, \highres{}, and all data.
We address our research questions in light of the results:






\begin{table}[!tb]
    \begin{subtable}[t!]{0.5\textwidth}
\centering
\small
\resizebox{\linewidth}{!}{
\begin{tabular}{llcccccc}
\toprule
        & \textbf{Lang. set}  & \multicolumn{2}{c}{\textbf{HL}} & \multicolumn{2}{c}{\textbf{HL balanced}} & \multicolumn{2}{c}{\textbf{Ours}} \\
        &     & I-L & Z-S &      I-L & Z-S & I-L & Z-S \\
\midrule
\multirow{4}{*}{\rotatebox[origin=c]{90}{\textbf{Shared}}} & Low-Res &    35.2 &      33.4 &         35.5 &      34.3 &    \textbf{36.6} &      \textbf{34.5} \\
        & High-Res &    83.3 &      33.7 &         81.2 &      32.4 &    \textbf{84.3} &      \textbf{33.8} \\
        & \{de\} &  \bf87.1  & 32.3  & 84.1 &
        32.2  & 86.8 & \bf33.0 \\
        & All &    55.8 &      33.5 &         55.1 &      33.5 &    \textbf{57.0} &      \textbf{34.2} \\
\midrule
\multirow{4}{*}{\rotatebox[origin=c]{90}{\textbf{Diverse}}} & Low-Res &    53.1 &      35.8 &         54.6 &      34.9 &    \textbf{55.7} &      \textbf{35.9} \\
        & High-Res &    76.8 &      36.2 &         73.4 &      34.7 &    \textbf{77.3} &      \textbf{36.8} \\
        & \{de\} &  \bf87.7  & 36.8  & 83.3 &
        35.3  & 87.4 & \bf38.1 \\
        & All &    63.3 &      36.0 &         62.7 &      34.8 &    \textbf{64.9} &      \textbf{36.3} \\
\bottomrule
\end{tabular}
}
\caption{Accuracy of POS probing.}
\label{tab:pos-res}
\end{subtable}

    \hfill
    \begin{subtable}[!t]{0.5\textwidth}
\centering
\small
\resizebox{\linewidth}{!}{
\begin{tabular}{llcccccc}
\toprule
        & \textbf{Lang. set}  & \multicolumn{2}{c}{\textbf{HL}} & \multicolumn{2}{c}{\textbf{HL balanced}} & \multicolumn{2}{c}{\textbf{Ours}} \\
        &     & I-L & Z-S &      I-L & Z-S & I-L & Z-S \\
\midrule
\multirow{4}{*}{\rotatebox[origin=c]{90}{\textbf{Shared}}}& Low-Res &    26.5 &      23.7 &         27.9 &      \bf24.3 &   \bf29.8 &      23.9 \\
        & High-Res &    34.2 &      24.9 &         34.7 &      24.7 &    \bf37.6 &      \bf26.0 \\
        & \{de\} &  31.4  & \bf27.4  & \bf32.1 &
        25.7  & 32.0 & 23.9 \\
        & All &    29.8 &      24.2 &         30.8 &      24.5 &    \bf33.1 &      \bf24.8 \\ \midrule
\multirow{4}{*}{\rotatebox[origin=c]{90}{\textbf{Diverse}}} & Low-Res &    25.7 &      12.8 &         28.0 &      \bf13.8 &    \bf29.9 &      12.9 \\
        & High-Res &    32.8 &      14.9 &         29.9 &      15.1 &    \bf37.2 &      \bf17.1 \\
        & \{de\} &  32.5  & 14.8  & 31.5 &
        15.7  & \bf35.3 & \bf17.2 \\
        & All &    28.7 &      13.7 &         28.8 &      14.4 &    \bf33.0 &      \bf14.7 \\
\bottomrule
\end{tabular}
}
\caption{Macro F1 of NER probing.}
\label{tab:ner-results}
\end{subtable}

    \caption{For each model and language set, we report average \inlang{} performance (probe trained and tested on the same language) and average \zeroshot{} performance (probe trained on one language and tested on another). Each \zeroshot{} number is an average result across all source languages and target languages in the indicated language set.  Each entry is a mean of 5 runs with different probe initialization.
    The Results with significance intervals for each language can be found in Appendix (Tables~ \ref{tab:pos-res-per-lang}, \ref{tab:ner-results-all-langs}).}
    \label{tab:probing-res}
\end{table}

\paragraph{Our method offers a good trade-off between performance on high- and low-resource languages.}

Figure~\ref{fig:lm-results} shows the trend of language modeling scores (MRR) when changing from low- to \highres{} set.
Table~\ref{tab:dist_from_mono} summarizes performance differences from monolingual models for our method and the two control baseline models. 

In \lowres{} setting, our model outperforms \ceunbalanced{} and achieves similar results to \cebalanced{}.
For \highres{} languages, our approach closely trails \ceunbalanced{} and is better than \cebalanced{}, which was trained on the same data as our student model. It indicates that the student model effectively acquires knowledge from the teachers' distributions.
Our model achieves the best results overall when calculated over all languages.

\paragraph{Better trade-off between high- and low-resource languages improves results on downstream.} 
\label{sec:cross-lingual-transfer}
Table~\ref{tab:probing-res} shows that \inlang{} and \zeroshot{} results of probing for POS and NER labels. Our method achieves better or on-par average results in both tasks and language sets. The only exception is \cebalanced{} baselines, which scores better in NER for \lowres{} languages.

\paragraph{Sharing script is not necessary for good multilingual performance.}

As seen in Figure~\ref{fig:lm-results} and Table~\ref{tab:dist_from_mono} for \lowres{} languages, shared script results are consistently closer to monolingual results compared to the diverse script setting. Whereas, for \highres{} set, the average difference between the results of monolingual models and our model or \ceunbalanced{} is smaller in the diverse script scenario.
For the language included in both sets (German),  MRR is higher when coupled with distinct script languages. The performance difference is 0.4 and 0.9 percent in favor of diverse scripts, for \ceunbalanced{} and our model. \cebalanced{} scores 2.8\% better in shared script scenario. This implies that diverse scripts can benefit multilingual modeling when we reveal enough monolingual data (as in \highres{} setting).

In Table~\ref{tab:probing-res}, we observe that the results for German in the shared-script scenario are better for POS tagging and worse for NER in comparison to diverse-script.
Those findings align with previous results suggesting that shared vocabulary is not necessary for cross-lingual transfer and has a varying effect depending on the task~\citep{wang2019cross, malkin2022balanced}.


\section{Related Work}
\label{sec:related-work}
Recent work utilized knowledge distillation in training NLP models. However, to the best of our knowledge, we are the first to do this in low-resource, balanced data settings. 
Contrary to the approaches of \citet{tsai2019small, sanh2019distilbert}, we do not scale down student models but constraint training datasets. 

\citet{sun2020knowledge} use one teacher model and train for machine translation, and
\citet{heffernan2022bitext} use a single multilingual teacher to train a sentence embedding model for low-resource languages. Both rely on parallel corpora for target low-resource languages.
Other works on multilingual language modeling addressed how to improve low-resource performance, largely using post-hoc or language-specific solutions. \citet{chau2020parsing} change the vocabulary to account for low-resource languages, while \citet{muller2020being} transliterate tokens of low-resource languages to the most similar available high-resource language.

Finally, \citet{pfeiffer2020mad} introduce cross-lingual adapters, compact components that allow adapting a given model pre-trained for a task in a different desired language. 

\section{Conclusions}
\label{sec:conclusions}
We train multilingual language models aimed at balancing the models' performance for languages with uneven data sizes. We outperform standard models for low-resource languages while maintaining performance on high-resource languages. 
Noticeably, our method gives better results overall than the naive data sub-sampling. 
Lastly, our model is a good representation learner for downstream tasks, outperforming baselines for two  probing tasks. 

Taken together, our results suggest a new direction for multilingual modeling that accounts for a more even performance across low- and high-resource languages and improves cross-lingual transfer.


\section{Limitations}
\label{sec:limitations}

\paragraph{Restricted model size and training.}
Due to limited computational resources, we performed experiments for models significantly smaller than the ones developed by the industry. We based our down-scaling choices on previous ablation studies on cross-lingual models \citep{wang2019cross}. In line with their findings, we prioritized model depth (6 hidden layers) over width (4 attention heads). Also, we examine only \textsc{BERT} based models. This work serves as a proof of concept for a new multilingual language modeling, and future work can extend the study to bigger models with different architectures. 

\paragraph{Restricted data.} We decided to train our models on sub-sampled Wikipedia to achieve reasonable training times. As shown in \cref{sec:datasize-our-vs-wiki} the chosen sample follows the resource-richness trend across languages but does not fully reflect the imbalance between high- and low-resource languages. Nevertheless, we think that this issue does not weaken our point, as even our ``unbalanced'' baseline model is trained on less skewed data than currently deployed multilingual models. Furthermore, we train our models on 7 languages. Our method needs to be verified on larger data sizes and broader language sets.

Working with limited training data might still be valuable in several aspects. First, there’s a growing interest in efficient, and green AI. Smaller and more efficient models will reduce training and inference costs while allowing them to run on less capable hardware and make them accessible to a wider community. Second, from a linguistic perspective, many of the world’s languages lack large corpora, and hence will benefit from models that leverage a limited amount of available resources \cite{joshi-etal-2020-state}.

\paragraph{Naive balancing method.} We truncate our training to the size of the smallest low-resource languages, which might be a naive and aggressive approach leading to a sub-optimal performance on our available data. However, our simple approach achieves good results even with naive balancing. Future work can extend it with complex data balancing approaches, such as weighing training data using a learned data scorer (as done in \citet{wang2020balancing}). 

\label{teacher_availbility}
\paragraph{Teacher model availability.} Our teacher-student training method assumes the existence of pre-trained monolingual teachers for each considered language, which is considerably less sustainable than training only one multilingual model. Nevertheless, we believe that it is possible to re-use publicly available models as teachers for high-resource languages, while for low-resource languages, competitive results can be obtained with smaller models requiring less computation \citep{hoffmann-borgeaud-mensch2022optimal}.
Because our distillation method works on predicted distribution and not latent representations, to combine knowledge of teachers from multiple source languages, we will need to align their vocabularies, which was shown to be feasible by \citet{artetxe-etal-2020-cross,rust-etal-2021-good}. We leave this  engineering task for future work.

\paragraph{Metrics for probing tasks.} To evaluate probing for NER we used macro-F1 measured per token and not per entity as in usual NER evaluation. We observed that the probes underperformed in correctly classifying all tokens in a single entity. It led to overall low results in regular F1 that would not allow meaningful comparison between analyzed models. Importantly, macro-F1 equally weights the performance in predicting each class. Thus, it is appropriate to evaluate NER task, where most tokens are annotated as not belonging to any entity.

\section*{Acknowledgements}
We thank anonymous reviewers for their valuable comments on the previous versions of this article. This work was supported in part by a research gift from the Allen Institute for AI, and a research grant 2336 from the Israeli Ministry of Science and Technology. Tomasz Limisiewicz's visit to the Hebrew University has been supported by grant 338521 of the Charles University Grant Agency and the Mobility Fund of Charles University.

\bibliography{custom}
\bibliographystyle{acl_natbib}
\clearpage

\appendix
\section{Appendix}
\label{sec:appendix}
In the appendix: we provide details on datasets used in this work Section~\ref{sec:data-details}; show how proposed teacher-student distillation behaves in the monolingual scenario with just one teacher Section~\ref{sec:TS-monolingual}; present detailed results of our two experimental for each language Section~\ref{sec:per-lang-results}; provide details of our training procedure and hardware usage Section~\ref{sec:reproduce}.

\section{Datasets Details}
\label{sec:data-details}

\subsection{Data Splits}
\begin{table}[!tb]
\centering
\begin{tabular}{@{}lllll@{}}
\toprule
\multirow{2}{*}{Lang.} & \multicolumn{2}{c}{POS} & \multicolumn{2}{c}{NER} \\ 
                       & train       & test      & train      & test       \\ \midrule
de                     & 166849      & 22458     & 20000      & 10000      \\
es                     & 28492       & 3147      & 20000      & 10000      \\
en                     & 21253       & 5440      & 20000      & 10000      \\
eu                     & 5396        & 1799      & 10000      & 10000      \\
hu                     & 910         & 449       & 20000      & 10000      \\
tr                     & 3664        & 4785      & 20000      & 10000      \\
vi                     & 1400        & 800       & 20000      & 10000      \\ \midrule
ru                     & 67435       & 11336     & 20000      & 10000      \\
ko                     & 27410       & 4276      & 20000      & 10000      \\
el                     & 28152       & 2809      & 20000      & 10000      \\
hi                     & 13304       & 2684      & 5000       & 1000       \\
te                     & 1051        & 146       & 1000       & 1000       \\
ur                     & 4043        & 535       & 20000      & 1000       \\ \bottomrule
\end{tabular}
\caption{Number of training and testing sentences for POS and NER tasks in \xtreme{} data collection. The data were used to train and evaluate probes on top of analysed models.}
\label{tab:train-test-split}
\end{table}

For pre-training (monolingual) teacher and \ceunbalanced{} models, we use Wikipedia splits of sizes indicated in Table~\ref{tab:lang}, for training student and \cebalanced{} models, we subsample training corpus to 10 million characters. We use validation and test sets containing 10000 Wikipedia sentences each.

For downstream probing, we use train and test splits from \xtreme{}. The numbers of sentences in these splits per language are shown in table~\ref{tab:train-test-split}.

\subsection{Correspondence of the Sizes of Our Corpora and Wikipedias}
\label{sec:datasize-our-vs-wiki}

\begin{figure}[t!]
    \centering
    \includegraphics[width=\linewidth]{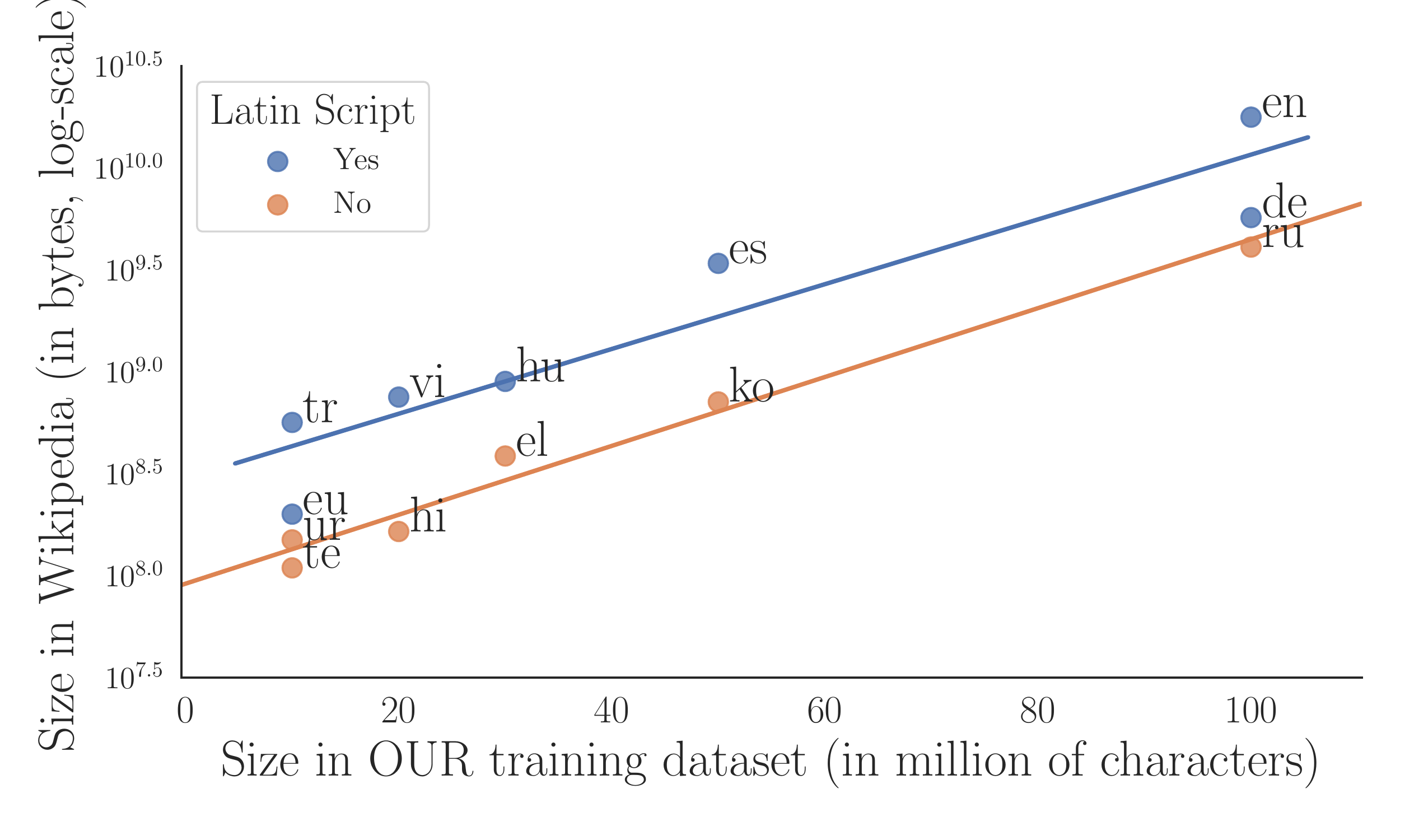}
    \caption{
    A comparison of subsampled corpora sized and the data available in Wikipedia, which was \textsc{mBERT}'s training corpus.   \label{fig:lang_choice_wiki}}
\end{figure}

Figure ~\ref{fig:lang_choice_wiki} shows the per language correspondence between our corpora size and the whole Wikipedia. The latter was used to pre-train \mbert{}~\citep{devlin-etal-2019-bert}. We observe a good linear fit between character numbers in our corpora and the logarithm of Wikipedia byte size. It suggests that the multilingual imbalance is even more severe in the original dataset than in our sample.

\section{Teacher-Student Method in the Monolingual World}
\label{sec:TS-monolingual}

\begin{figure}[tb!]
    \centering
    \includegraphics[width=\linewidth]{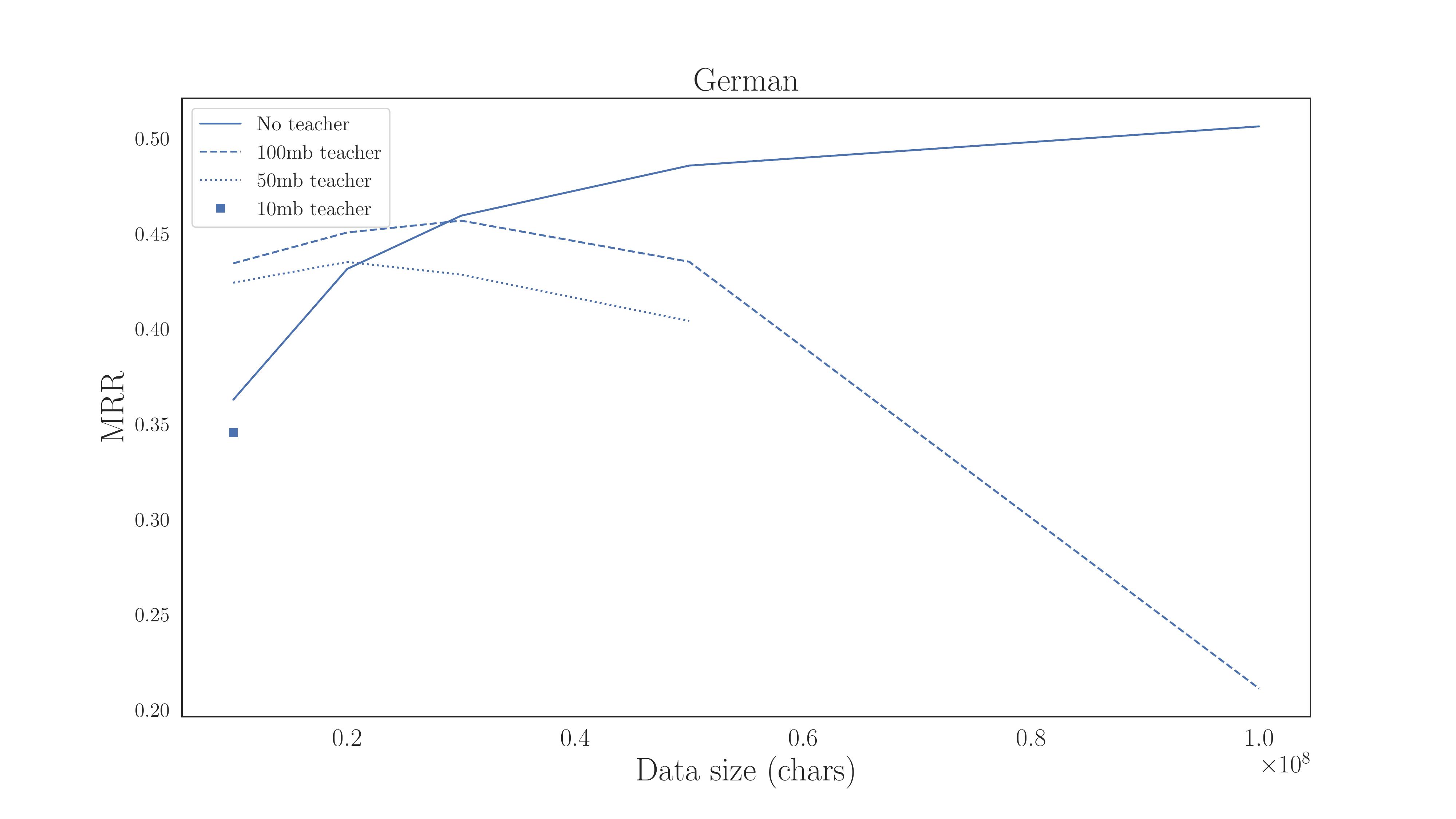}
    \caption{Performance of a language model as the function of training corpora size. The regular HL training is compared with the knowledge distillation to a student on the dataset lower or equal in size than the teacher's training set.}
    \label{fig:monolingual-ts}
\end{figure}

The purpose of this experiment is to visualize how the model's performance scales with the size of the pre-training dataset. Also, we check the behavior of the teacher-student knowledge distillation with the change of data size used to train a teacher and a student in a monolingual setting. 

We train a monolingual model on German Wikipedia data with five sizes (in millions of characters): 10, 20, 30, 50, and 100. Subsequently, we designate 10, 50, and 100 million character models as teachers and distill their knowledge into students on the same size or smaller corpus.\footnote{In monolingual knowledge distillation, we used a learning rate 5 times higher than in the default \textsc{BERT} training script. This choice led to better results.}

As presented in figure~\ref{fig:monolingual-ts}, the teacher performance can be nearly matched by a student trained on a considerably smaller corpus. For the teacher trained on the largest split, the student performance rises steadily with the increase of distillation detest from 10 to 30 million characters and drops after that point. The performance of the student trained on 100 million characters is noticeably low. It is a sign of over-fitting, as in our setting, distillation set is always a subset of the teacher's training set. Also, in the case of teachers trained on smaller corpora, distillation on the dataset of the same size (as the teacher training set) leads to a drop in performance. Therefore, we claim that the distillation is beneficial when the teacher's training set is larger than the student's one.

\section{Per Language Results}
\label{sec:per-lang-results}
\subsection{German: Comparing Shared and Diverse Scripts}
\begin{table}[t!]
\centering
\small
\resizebox{\columnwidth}{!}{
\begin{tabular}{lll}
\hline
\textbf{}            & \textbf{Shared script} & \textbf{Diverse script} \\ \hline
\textbf{HL}          & -2.9                   & \textbf{-2.5}           \\
\textbf{HL Balanced} & \textbf{-9.2}          & -12                     \\
\textbf{Ours}        & -6.1                   & \textbf{-5.2}           \\ \hline
\end{tabular}
}
\caption{Difference from monolingual baseline, for German. German achieves better results in diverse script, except for \emph{HL Balanced}. This suggest that diverse script might help increase language modeling performance.}
\label{tab:dist_from_mono_de}
\end{table}

\begin{figure}[tb!]
    \centering
    \includegraphics[width=\linewidth]{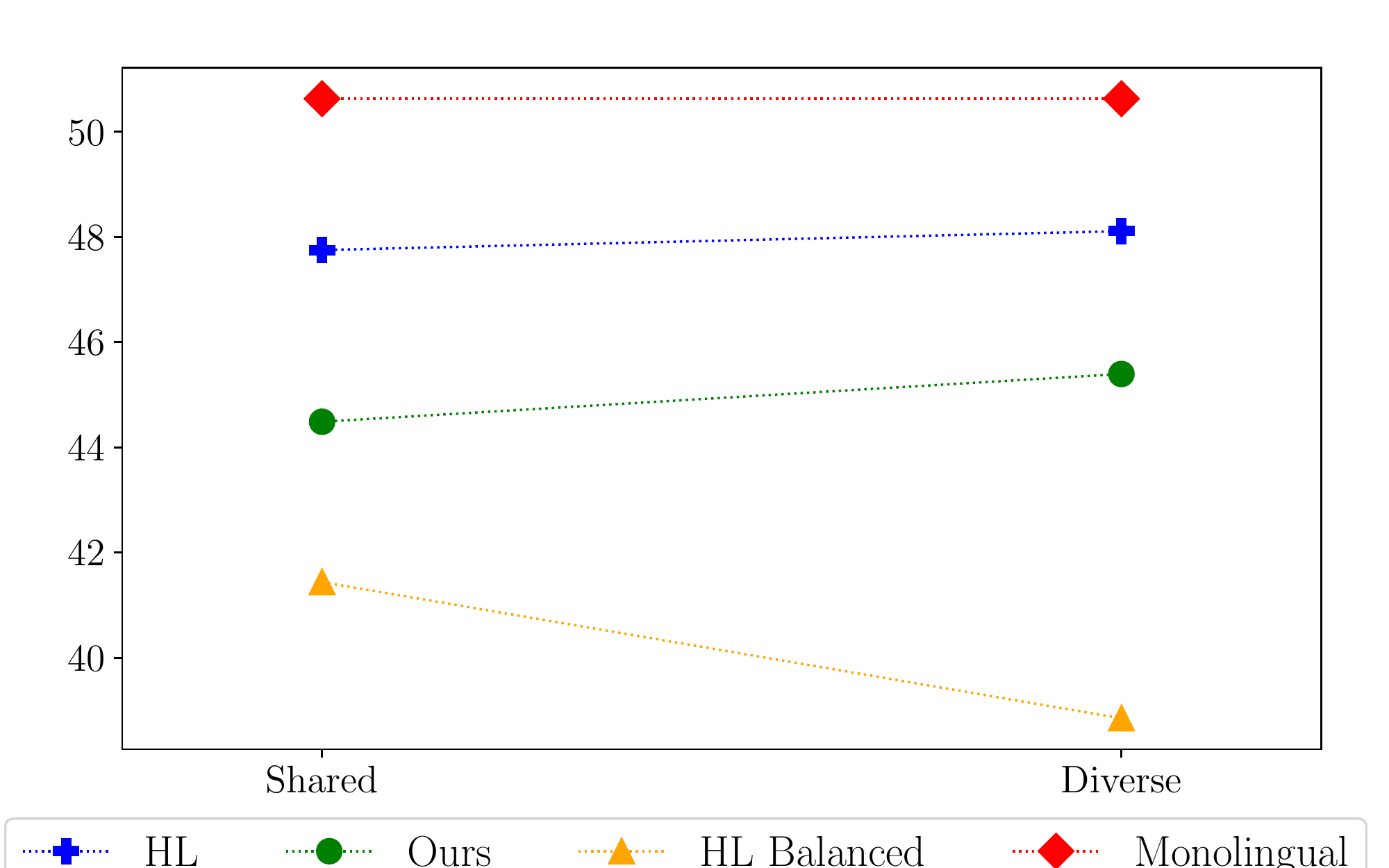}
    \caption{MRR scores for German trained in the set of languages with shared script and diverse script. We observe slight improvement for diverse script over shared script, and significant deterioration for \emph{HL Balanced}.}
    \label{fig:lm-results-german}
\end{figure}

Table~\ref{tab:dist_from_mono_de} and Figure~\ref{fig:lm-results-german} present masked language modeling performance for German for three analyzed multilingual model types. German is the language included both in the  shared and diverse script language sets. Therefore the results allow comparing which setting is more effective in multilingual language modeling. 

\subsection{Results for Every Language}

\begin{figure*}[tb!]
    \centering
    \begin{subfigure}[b]{\linewidth}
        \includegraphics[width=\linewidth]{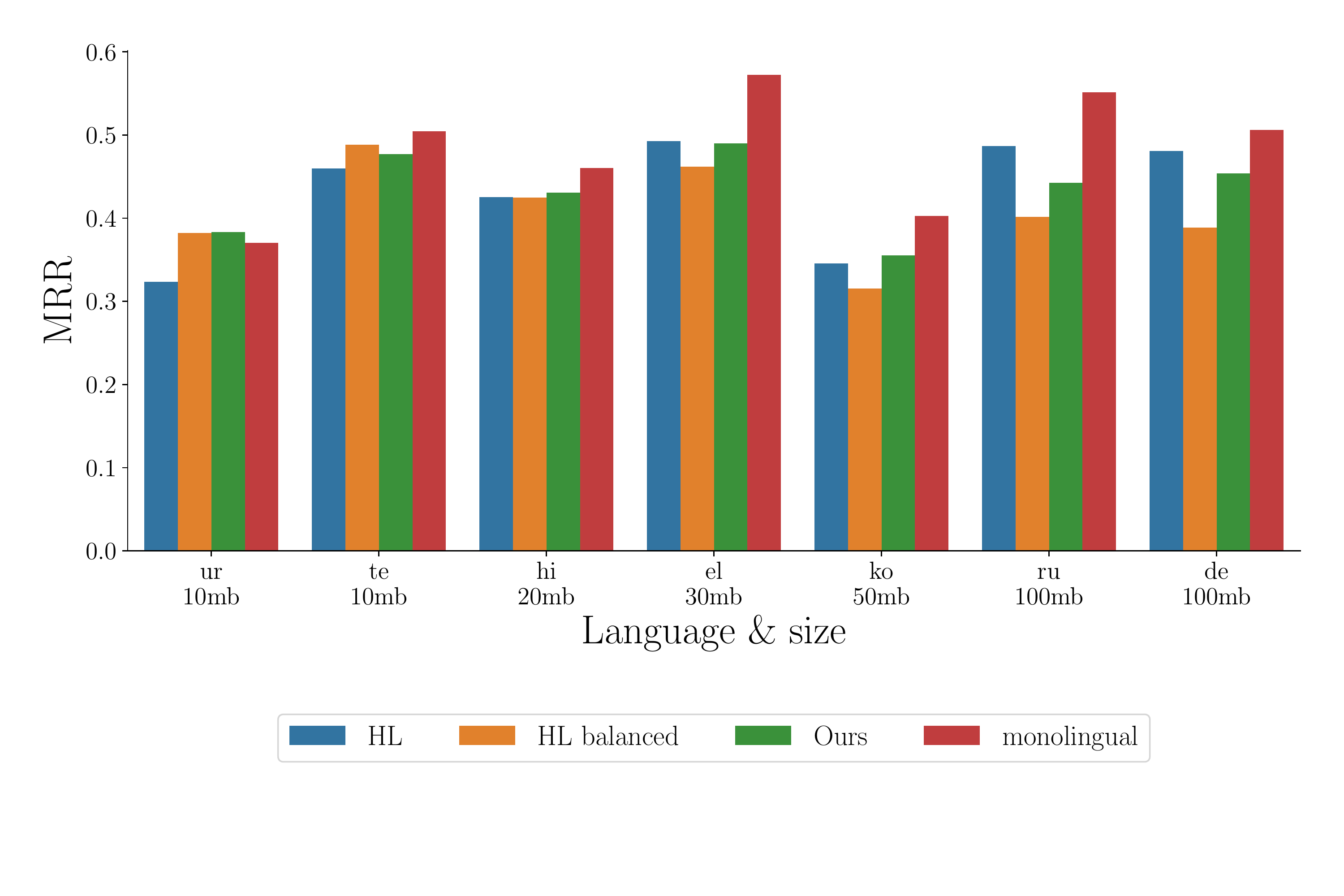}
        \caption{Shared Script (Latin)}
    \end{subfigure}%
    \vfill%
    \begin{subfigure}[b]{\linewidth}
        \includegraphics[width=\linewidth]{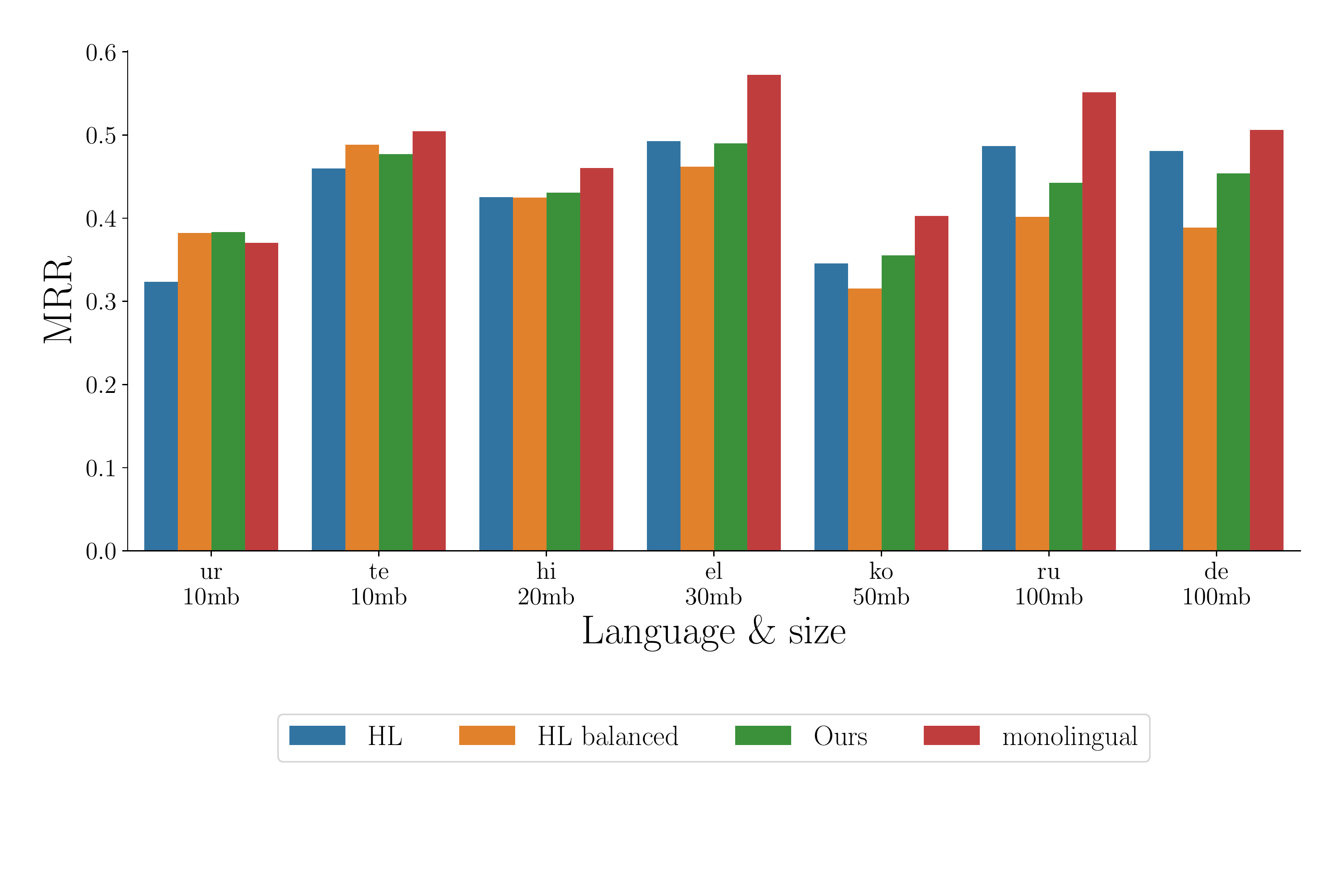}
        \caption{Diverse Script}
    \end{subfigure}
    \caption{The figures present MRR results for each language. Our  model is compared with baselines: \cebalanced{}, \ceunbalanced{} and monolingual models. We observe similar trends as in Figure~\ref{fig:lm-results} at higher granularity.}
    \label{fig:lm-results-per-lang}
\end{figure*}

We present per language results in masked language modeling performance in Figure~\ref{fig:lm-results-per-lang} and for probing tasks (POS and NER) in Tables~\ref{tab:pos-res-per-lang}~and~\ref{tab:ner-results-all-langs}.

\begin{table*}[t!]
\centering
\small
\begin{tabular}{lc|cc|cc|cc}
\toprule
Script & Lang. & \multicolumn{2}{c|}{HL} & \multicolumn{2}{c|}{HL balanced} & \multicolumn{2}{c}{Ours} \\
               &     &           In-Lang &         Zero-Shot &           In-Lang &         Zero-Shot &           In-Lang &         Zero-Shot \\
\midrule
\multirow{7}{*}{Shared} & de &  87.1 $_{\pm0.0}$ &  32.3 $_{\pm0.9}$ &  84.1 $_{\pm0.0}$ &  32.2 $_{\pm1.0}$ &  86.8 $_{\pm0.0}$ &  33.0 $_{\pm1.1}$ \\
        & en &  79.5 $_{\pm0.1}$ &  34.2 $_{\pm1.4}$ &  77.4 $_{\pm0.2}$ &  32.1 $_{\pm2.1}$ &  81.1 $_{\pm0.2}$ &  34.1 $_{\pm1.3}$ \\
        & es &  83.1 $_{\pm0.1}$ &  34.6 $_{\pm1.7}$ &  82.0 $_{\pm0.1}$ &  32.8 $_{\pm1.7}$ &  84.8 $_{\pm0.1}$ &  34.2 $_{\pm1.0}$ \\
        & eu &  56.3 $_{\pm1.2}$ &  34.1 $_{\pm1.2}$ &  58.1 $_{\pm1.5}$ &  35.0 $_{\pm2.7}$ &  58.2 $_{\pm0.7}$ &  33.1 $_{\pm2.2}$ \\
        & hu &  18.5 $_{\pm3.5}$ &  37.4 $_{\pm1.0}$ &  16.6 $_{\pm3.4}$ &  37.9 $_{\pm1.7}$ &  18.5 $_{\pm5.2}$ &  39.5 $_{\pm1.2}$ \\
        & tr &  40.5 $_{\pm2.2}$ &  33.3 $_{\pm1.5}$ &  40.6 $_{\pm3.8}$ &  34.5 $_{\pm2.2}$ &  42.1 $_{\pm2.6}$ &  34.1 $_{\pm2.6}$ \\
        & vi &  25.5 $_{\pm2.2}$ &  28.7 $_{\pm1.1}$ &  26.9 $_{\pm3.1}$ &  29.9 $_{\pm1.3}$ &  27.7 $_{\pm4.5}$ &  31.3 $_{\pm1.9}$ \\
\cline{1-8}
\multirow{7}{*}{Diverse} & de &  87.7 $_{\pm0.0}$ &  36.8 $_{\pm0.9}$ &  83.3 $_{\pm0.0}$ &  35.3 $_{\pm1.1}$ &  87.4 $_{\pm0.0}$ &  38.1 $_{\pm0.3}$ \\
        & ru &  79.0 $_{\pm0.0}$ &  36.9 $_{\pm0.9}$ &  74.0 $_{\pm0.1}$ &  36.9 $_{\pm1.2}$ &  78.6 $_{\pm0.1}$ &  38.8 $_{\pm1.5}$ \\
        & ko &  63.7 $_{\pm0.2}$ &  34.8 $_{\pm1.6}$ &  62.8 $_{\pm0.2}$ &  31.9 $_{\pm1.8}$ &  65.8 $_{\pm0.2}$ &  33.5 $_{\pm1.2}$ \\
        & el &  66.6 $_{\pm0.2}$ &  29.9 $_{\pm1.1}$ &  66.7 $_{\pm0.2}$ &  27.0 $_{\pm1.0}$ &  69.0 $_{\pm0.3}$ &  30.8 $_{\pm1.4}$ \\
        & hi &  70.7 $_{\pm0.2}$ &  34.9 $_{\pm0.8}$ &  69.6 $_{\pm0.1}$ &  34.7 $_{\pm1.9}$ &  70.8 $_{\pm0.4}$ &  36.0 $_{\pm1.6}$ \\
        & te &  25.1 $_{\pm9.9}$ &  42.1 $_{\pm1.8}$ &  30.0 $_{\pm8.4}$ &  43.0 $_{\pm1.2}$ &  28.4 $_{\pm6.0}$ &  42.6 $_{\pm1.3}$ \\
        & ur &  50.0 $_{\pm1.0}$ &  36.3 $_{\pm0.6}$ &  52.2 $_{\pm3.4}$ &  34.7 $_{\pm0.8}$ &  54.4 $_{\pm2.0}$ &  34.1 $_{\pm1.4}$ \\
\bottomrule
\end{tabular}
\caption{Accuracy of POS probing for each language. Standard deviations and mean results are computed based on 5 runs with different initialization of the probe.}
\label{tab:pos-res-per-lang}
\end{table*}

\begin{table*}[t!]
\centering
\small
\begin{tabular}{lc|cc|cc|cc}
\toprule
Script & Lang. & \multicolumn{2}{c|}{HL} & \multicolumn{2}{c|}{HL balanced} & \multicolumn{2}{c}{Ours} \\
               &     &           In-Lang &         Zero-Shot &           In-Lang &         Zero-Shot &           In-Lang &         Zero-Shot \\
\midrule
\multirow{7}{*}{Shared} & de &  31.4 $_{\pm0.6}$ &  27.4 $_{\pm1.0}$ &  32.1 $_{\pm0.4}$ &  25.7 $_{\pm0.4}$ &  32.0 $_{\pm0.7}$ &  26.9 $_{\pm1.1}$ \\
        & en &  33.0 $_{\pm0.5}$ &  24.9 $_{\pm0.7}$ &  33.3 $_{\pm0.4}$ &  24.8 $_{\pm0.2}$ &  37.8 $_{\pm0.7}$ &  25.9 $_{\pm0.9}$ \\
        & es &  38.2 $_{\pm0.6}$ &  22.6 $_{\pm0.3}$ &  38.8 $_{\pm1.3}$ &  23.7 $_{\pm0.7}$ &  42.9 $_{\pm1.0}$ &  25.4 $_{\pm1.7}$ \\
        & eu &  20.6 $_{\pm2.0}$ &  27.3 $_{\pm0.9}$ &  18.5 $_{\pm1.3}$ &  27.9 $_{\pm0.9}$ &  20.5 $_{\pm0.9}$ &  25.9 $_{\pm1.0}$ \\
        & hu &  26.6 $_{\pm0.5}$ &  24.3 $_{\pm0.9}$ &  26.8 $_{\pm1.0}$ &  25.3 $_{\pm0.4}$ &  30.2 $_{\pm0.6}$ &  26.1 $_{\pm0.9}$ \\
        & tr &  27.3 $_{\pm0.5}$ &  24.4 $_{\pm1.0}$ &  30.8 $_{\pm0.5}$ &  25.4 $_{\pm0.4}$ &  29.5 $_{\pm0.4}$ &  24.2 $_{\pm0.4}$ \\
        & vi &  31.5 $_{\pm1.4}$ &  18.7 $_{\pm0.5}$ &  35.5 $_{\pm0.5}$ &  18.6 $_{\pm0.5}$ &  39.0 $_{\pm1.5}$ &  19.2 $_{\pm1.3}$ \\ \midrule
\multirow{7}{*}{Diverse} & de &  32.5 $_{\pm0.8}$ &  14.8 $_{\pm0.6}$ &  31.5 $_{\pm0.7}$ &  15.7 $_{\pm0.7}$ &  35.3 $_{\pm0.4}$ &  17.2 $_{\pm1.0}$ \\
        & ru &  33.7 $_{\pm0.8}$ &  15.8 $_{\pm0.7}$ &  29.9 $_{\pm0.7}$ &  14.6 $_{\pm0.7}$ &  38.0 $_{\pm0.2}$ &  16.8 $_{\pm0.7}$ \\
        & ko &  32.1 $_{\pm0.4}$ &  14.2 $_{\pm0.4}$ &  28.2 $_{\pm0.5}$ &  15.0 $_{\pm0.4}$ &  38.3 $_{\pm0.8}$ &  17.3 $_{\pm1.1}$ \\
        & el &  27.4 $_{\pm0.7}$ &  16.6 $_{\pm0.6}$ &  26.5 $_{\pm0.9}$ &  17.2 $_{\pm0.8}$ &  31.5 $_{\pm0.6}$ &  16.6 $_{\pm0.6}$ \\
        & hi &  16.3 $_{\pm0.7}$ &  12.8 $_{\pm0.4}$ &  18.1 $_{\pm1.0}$ &  14.4 $_{\pm1.2}$ &  15.7 $_{\pm1.1}$ &  13.2 $_{\pm0.7}$ \\
        & te &  13.3 $_{\pm1.1}$ &  13.8 $_{\pm0.5}$ &  14.6 $_{\pm2.0}$ &  13.7 $_{\pm0.4}$ &  14.2 $_{\pm0.6}$ &  13.9 $_{\pm0.2}$ \\
        & ur &  45.6 $_{\pm1.1}$ &   7.9 $_{\pm1.4}$ &  52.7 $_{\pm1.2}$ &  10.0 $_{\pm1.2}$ &  58.0 $_{\pm1.0}$ &   8.0 $_{\pm0.9}$ \\
\bottomrule
\end{tabular}
\caption{Macro-F1 of NER probing for each language. Standard deviations and mean results are computed based on 5 runs with different initialization of the probe.}
\label{tab:ner-results-all-langs}
\end{table*}






\label{sec:app-ds-data}

\section{GPUs and training procedures}
\label{sec:reproduce}
All of our models (monolingual teachers, students, and multilingual models trained using \emph{hard labels}) are trained on a single GPU core. 

We used varying GPUs architectures allocated for each model upon availability (nvidia gtx 980, tesla M60, and RTX 2080Ti). 
Training time varied between 1 to 3 hours for monolingual models (depending on the data size, language, and GPU core). Multilingual models' training took around 18 hours to complete. Early stopping was used for all models based on results on a balanced dev set. 

MLM evaluation was run on the same machines as training or on CPU. the run time  ranged from 2 to 4 hours. Training a probe on top of a frozen model took from 1 to 20 minutes, depending on the number of training examples available for a language. The evaluation time on a downstream task was less than 2 minutes.

\end{document}